# Consumer-to-Clinical Language Shifts in Ambient AI Draft Notes and Clinician-Finalized Documentation: A Multi-level Analysis

Ha Na Cho, MS[1], Yawen Guo, MISM[1], Sairam Sutari MS[2], Emilie Chow MD[3], Steven Tam MD[3], Danielle Perret MD[4], Deepti Pandita MD[3], Kai Zheng, PhD[1]

[1]Department of Informatics, University of California, Irvine, Irvine, CA, USA;

[2]Institute for Clinical and Translational Science, University of California, Irvine, Irvine, CA, USA;

[3]Department of Medicine, University of California, Irvine, Irvine, CA, USA;

[4]Department of Physical Medicine & Rehabilitation, University of California, Irvine, Irvine, CA, USA

**Abstract**

*Ambient AI generates draft clinical notes from patient-clinician conversations, often using lay or consumer-oriented phrasing to support patient understanding instead of standardized clinical terminology. How clinicians revise these drafts for professional documentation conventions remains unclear. We quantified clinician editing for consumer-to-clinical normalization using a dictionary-confirmed transformation framework. We analyzed 71,173 AI-draft and finalized-note section pairs from 34,726 encounters. Confirmed transformations were defined as replacing a consumer expression with its dictionary-mapped clinical equivalent in the same section. Editing significantly reduced terminology density across all sections (p < 0.001). The Assessment and Plan accounted for the largest transformation volume (59.3%). Our analysis identified 7,576 transformation events across 4,114 note sections (5.8%), representing 1.2% consumer-term deletions. Transformation intensity varied across individual clinicians (p < 0.001). Overall, clinician post-editing demonstrates consistent shifts from conversational phrasing toward standardized, section-appropriate clinical terminology, supporting section-aware ambient AI design.*

**Introduction**

Ambient artificial intelligence (AI) systems convert real-time patient-clinician dialogue into draft clinical notes, aiming to reduce documentation burden while preserving encounter content[1, 2]. Unlike structured templates, these systems generate free-text narratives grounded in conversational language, which introduce an important distinction from the templated notes, because conversational dialogue employs consumer-oriented, lay expressions that diverge systematically from the standardized professional vocabulary required in finalized clinical records[3, 4]. As patients increasingly access clinical notes through open-notes mandates, this linguistic gap carries implications for health literacy, patient trust, and documentation quality[5].

Prior evaluations of ambient AI documentation have focused primarily on efficiency outcomes, time-to-completion, user satisfaction, and workflow integration, while leaving the linguistic content of post-editing largely unexamined[6, 7]. Existing characterizations note that AI-generated drafts require substantial clinical editing, yet existing observations remain largely qualitative[7]. A parallel body of work in consumer health informatics has demonstrated systematic differences between lay health language and biomedical terminology, with the Consumer Health Vocabulary (CHV) providing a comprehensive mapping infrastructure linking patient-facing expressions to professional clinical concepts[8, 9]. How these linguistic distinctions manifest within real-world ambient AI workflows, where consumer language is algorithmically captured and then modified through clinician editing, has not been systematically quantified.

This study addresses that gap through multi-level analysis of 71,173 paired AI-draft and clinician-finalized note-sections from a large-scale ambient documentation deployment. We introduce a dictionary-confirmed transformation framework that identifies editing events in which a consumer-oriented expression is removed from an AI draft while its dictionary-mapped clinical equivalent is simultaneously introduced in the finalized note within the same note-section. This co-occurrence constraint distinguishes targeted lexical normalization from incidental textual change, enabling reproducible phrase-level measurement of consumer-to-professional terminology substitution at scale. We apply this framework across four section types, History of Present Illness (HPI), Assessment and Plan (A&P), Physical Exam, and Results, alongside corpus-level, note-level, and clinician-level analyses.

RQ1. Do clinician-edited notes demonstrate systematic shifts in consumer and clinical terminology compared with

AI-generated drafts?
RQ2. To what extent do clinicians perform dictionary-confirmed consumer-to-clinical term transformations within the same note-section?
RQ3. Does editing behavior vary across sections, clinicians and clinical specialties?

**Methods**
*Study Setting and Cohort*
This retrospective analysis examined pre- and post-editing versions of ambient AI documentation collected during a pilot deployment from the University of California, Irvine (UCI) Health (December 2023-April 2025). Clinicians accessed the Abridge tool through Epic Haiku and the web interface. Draft note sections were generated during encounters and reviewed, edited, and finalized within the electronic health record (EHR). Encounter audio was transcribed with patient notification at the point of care. Draft and finalized note text were preserved through EHR audit trails. Each NOTE_ID represented a single clinical note containing multiple structured sections. Text fragments sharing the same NOTE_ID and section label were concatenated into a paired draft-final section instance, defined as a NOTE_ID-section unit. Because a single encounter could generate one or more clinical notes, the number of note documents (NOTE_ID) exceeded the number of encounters. Analyses were therefore conducted at the note-section level, defined as the combination of NOTE_ID and section label with draft and finalized text aggregated within each unit. Analyses focused on four sections with distinct communicative roles: History of Present Illness (HPI), Assessment and Plan (A&P), Physical Exam, and Results. The final dataset comprised 71,173 note-sections from 34,726 unique encounters authored by 241 clinicians (22,664 patients; Table 1). The study was approved by Institutional Review Board #7234 (UCI), and all processing was conducted in a HIPAA-compliant AWS cloud environment. An overview of the study pipeline is illustrated in Figure 1.

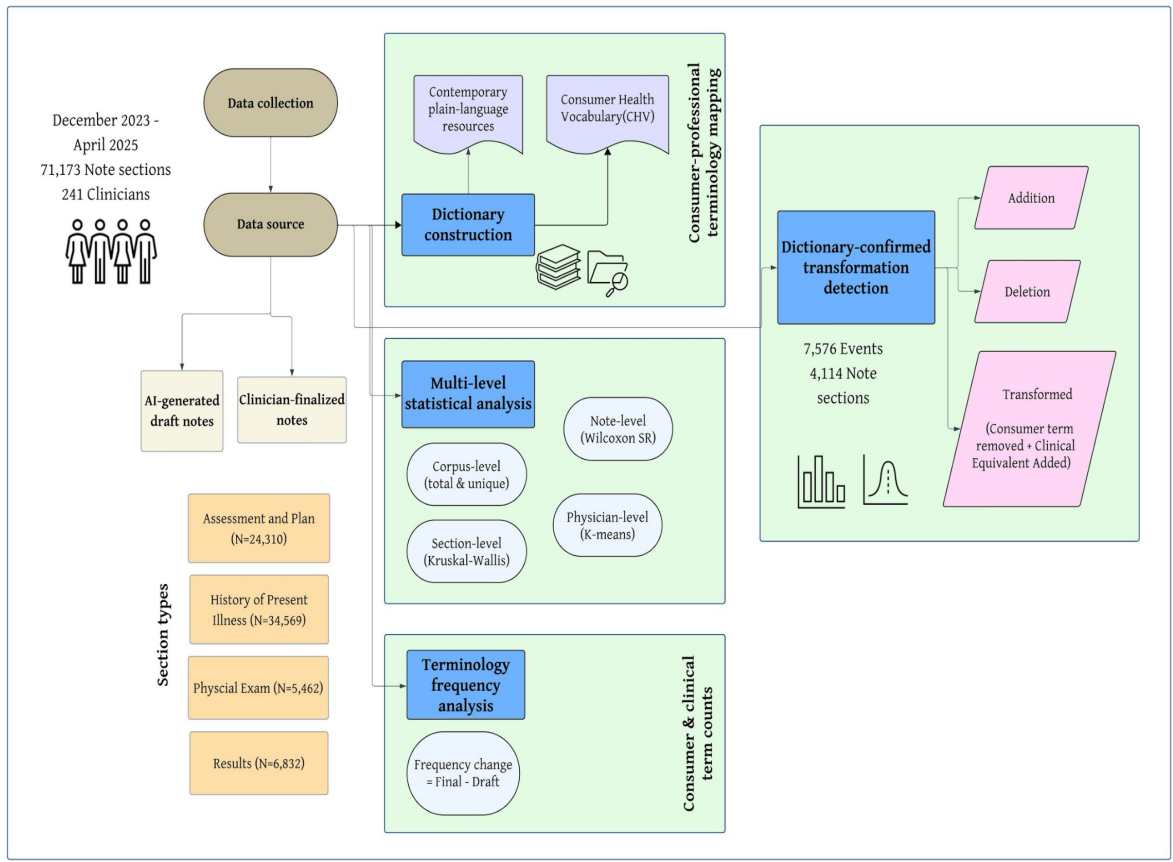

**Figure 1.** Pipeline for detecting dictionary-confirmed consumer-to-clinical terminology transformations in clinician-edited ambient AI documentation.

**Table 1.** Cohort characteristics.

|  | **Characteristic** | **n (%)** |
|---|---|---|
| **Note-sections** | Total note-sections analyzed | 71,173 |
|  | HPI | 34,569 (48.6) |
|  | A&P | 24,310 (34.2) |
|  | Results | 6,832 (9.6) |
|  | Physical Exam | 5,462 (7.7) |
| **Provider** | Medical Doctor (MD) | 159 (66.0) |
|  | Doctor of Osteopathic Medicine (DO) | 16 (6.6) |
|  | Nurse Practitioner (NP) | 11 (4.6) |
|  | Physician Assistant (PA) | 10 (4.1) |
|  | Optometric Doctor (OD) | 3 (1.2) |
|  | Unknown | 42 (17.4) |
| **Gender** | Female | 106 (44.0) |
|  | Male | 135 (56.0) |
| **Age** | Mean clinician age, years (SD) | 41.0 (10.4) |
|  | Mean patient age, years (SD) | 54.9 (18.9) |

*Dictionary Construction and Terminology Mapping*
A unified consumer-to-clinical terminology mapping dictionary combined the CHV, which provides UMLS-linked representation of lay expressions mapped to professional medical concepts[8-10], with contemporary plain-language terminology resources[11,12]. Entries were normalized through lowercase conversion and whitespace standardization, and CHV stop concepts and incorrect mappings were excluded[13]. Duplicates were removed based on normalized consumer-to-clinical term pairs. Although concept identifiers were retained for provenance tracking, downstream analyses were performed at the phrase level to support scalable text matching across a large corpus, consistent with prior consumer health informatics approaches emphasizing concept-level rather than direct lexical translation[14]. Phrases were retained only when all constituent tokens appeared in the corpus and when phrase length ranged from one to six tokens. Terminology identification was performed using case-insensitive, boundary-aware regular expressions to capture whole-word or whole-phrase occurrences without substring leakage, consistent with established clinical NLP preprocessing practices[15]. The final dictionary comprised 71,386 consumer-to-clinical mappings.

*Terminology Frequency Analysis*
Consumer-term and clinical-term frequencies were computed independently in draft and finalized text for each note-section using case-insensitive, boundary-aware regular expressions to prevent substring leakage. Within-section editing was quantified as change = final - draft frequency, capturing direction and magnitude of change. Analyses were conducted at corpus (aggregate), note (per-note), and section (per-section, stratified by type) levels.

*Dictionary-Confirmed Consumer-to-Clinical Transformation Detection*
A transformation event required two simultaneous conditions within a single note-section: (1) a consumer term present in the AI-draft is absent from the finalized text (removed); and (2) the clinical term to which that consumer term is dictionary-mapped is newly present in the finalized text and absent from the draft (added). Both conditions must hold within the same NOTE_ID-section unit, excluding incidental deletions, standalone insertions, and unmapped terminology changes. To efficiently detect dictionary-confirmed transformations across a large set of consumer-to-clinical mappings, terminology matching was performed using a two-stage strategy. Single-word terms were identified through token-level matching, whereas multi-word phrases were first narrowed to candidate occurrences using indexed keyword retrieval and then verified using boundary-aware regular expressions to ensure exact phrase matches. Terms shorter than four characters and high-frequency function words were excluded to minimize false positives. Detected transformations were summarized at two levels: event-level counts (per NOTE_ID-section instance) and pair-level counts (number of distinct note-sections in which a consumer-to-clinical pair occurred). A post-hoc linguistic filtering step removed same-stem morphological variants, short tokens, and domain-general stop words to retain clinically interpretable substitutions. Finally, a clinical relevance threshold of more than 10 note-sections was applied to identify reproducible transformation pairs for reporting.

*Clinician-Level Clustering*
Of 241 clinicians, 207 contributed to more than 10 note-sections, and were eligible for clustering. Unsupervised k-means clustering (k = 3) was applied to this subset using mean consumer-term changes and percentage zero-change sections as clustering variables. Editing profiles were characterized by mean terminology change, zero-change rate, and section volume per clinician.

*Statistical Analysis*
All draft-final comparisons were treated as paired observations. Two-sided Wilcoxon signed-rank tests evaluated terminology changes. Kruskal-Wallis tests compared clinician-level summaries across section types and specialty groups. Pairwise comparisons were conducted using Mann-Whitney U tests with Holm correction for multiple testing, with statistical significance defined as $p < 0.05$.

**Results**
*Cohort Characteristics*
The final dataset included 71,173 paired note-sections from 34,726 encounters (Table 1). HPI constituted the largest share (34,569; 48.6%), followed by A&P (24,310; 34.2%), Results (6,832; 9.6%), and Physical Exam (5,462; 7.7%). Among 241 clinicians, most held Doctor of Medicine degrees (159; 66.0%; mean age 41 years [SD 10.4]). Patients' mean age was 55 years (SD 18.9; range 18-104). Progress notes accounted for 97.3% of sections; office visits comprised 87.7%.

*Corpus-Level Shifts in Mapped Terminology*
Across 35,439 paired clinical notes derived from 34,726 encounters, clinician post-editing produced substantial reduction in both consumer and clinical terminology relative to AI-generated drafts (Table 2). Consumer-term total frequency decreased by 28.1% (3,814,042 to 2,742,428) and clinical-term total frequency by 29.1% (4,511,917 to 3,200,742), with modest parallel reductions in unique term diversity (consumer: 18,428 to 17,746, -3.7%; clinical: 13,633 to 13,338, -2.2%). These parallel reductions indicate that AI-generated drafts systematically over-generate both consumer and clinical terminology relative to finalized notes, with clinicians compressing overall lexical density during post-editing. At the note level (n = 35,439), Wilcoxon signed-rank tests confirmed uniform, highly significant reductions: median change in consumer = -18 ($p < 0.001$), median change in clinical = -23 ($p < 0.001$), median change in unique consumer = -7 ($p < 0.001$), median change in unique clinical = -9 ($p < 0.001$).

**Table 2. Corpus-level terminology frequency changes between draft and final notes.**

| Terminology type | Draft | Final | Change (%) |
| --- | --- | --- | --- |
| Consumer terms (total) | 3,814,042 | 2,742,428 | -1,071,614 (-28.1%) |
| Consumer terms (unique) | 18,428 | 17,746 | -682 (-3.7%) |
| Clinical terms (total) | 4,511,917 | 3,200,742 | -1,311,175 (-29.1%) |
| Clinical terms (unique) | 13,633 | 13,338 | -295 (-2.2%) |

*Section-Level Editing Patterns*
Section-level analysis revealed substantial heterogeneity in editing intensity across documentation components (Kruskal-Wallis $\chi^2$ = 2758.73, $p < 0.001$; Table 3). All four sections showed significant terminology reductions (Wilcoxon signed-rank, all $p < 0.001$), but magnitude and pattern differed markedly by communicative role. In the HPI section (34,569 sections), clinicians removed 648,066 consumer phrases and retained 841,335, corresponding to a removal proportion of approximately 43.5 percent and an editing intensity of 18.75 consumer-term deletions per note. Concurrently, 173,738 clinical terms were newly introduced while 943,309 were retained, averaging 5.02 clinical-term additions per note and indicating gradual expansion of professional terminology without wholesale replacement. The A&P section (24,310 sections) showed a distinct editing profile, with 321,968 consumer-term deletions and 604,765 retained consumer phrases, reflecting a removal proportion near 34.7 percent and approximately 13.24 deletions per note. This section also exhibited the highest level of clinical expansion, including 224,029 clinical-term additions or 9.21 added professional terms per note, suggesting active professional reframing of plan-related narratives. Editing intensity was substantially lower in more structured documentation areas. Results (6,832 sections) showed 35,940 consumer-term deletions relative to 13,192 retained phrases, producing a higher proportional removal despite only 5.26 deletions per note, while Physical Exam (5,462 sections) demonstrated 18,717 deletions or 3.43 per note alongside modest clinical-term additions. Across all sections, clinicians preserved a large base of existing clinical language while selectively removing conversational phrasing, indicating additive lexical normalization rather than wholesale rewriting.

Section-level differences in normalized editing intensity were statistically significant (Kruskal-Wallis test, $p < 0.001$), confirming that narrative sections such as HPI and A&P exhibited greater terminology normalization compared with more structured sections. Pairwise comparisons confirmed that HPI was significantly more intensively edited than Physical Exam (Mann-Whitney U, consumer median $p = 0.0004$, Holm-corrected $p = 0.002$). Differences between A&P and structured sections were directionally consistent but did not reach Holm-corrected significance, reflecting greater within-section variability in A&P narrative content.

**Table 3.** Section-level editing intensity and mean per-note terminology changes during clinician revision of AI-generated clinical documentation.

| Section | Consumer-term deleted (mean/note) | Consumer-term added (mean/note) | Clinical-term added (mean/note) | Consumer-term deletion (%) | Net consumer change | P value |
|---|---|---|---|---|---|---|
| HPI | 18.75 | 4.54 | 5.02 | 43.5 | -14.21 | <0.001 |
| A&P | 13.24 | 8.28 | 9.21 | 34.7 | -4.96 | <0.001 |
| Results | 5.26 | 1.45 | 1.58 | 73.1 | -3.81 | <0.001 |
| Physical Exam | 3.43 | 1.72 | 2.24 | 59.4 | -1.71 | <0.001 |

*Dictionary-Confirmed Consumer-to-Clinical Transformations*
Applying the co-occurrence transformation algorithm across 71,173 note-sections identified 7,576 confirmed transformation events in 4,114 note-sections (5.8%), representing 1,213 unique consumer-to-clinical pairs prior to linguistic filtering. Figure 2 illustrates a representative example from an A&P section, demonstrating co-occurring consumer-term deletions and mapped clinical-term insertion within the same note-section. The example shown uses deidentified synthetic text for illustrative purposes. After removing same-stem morphological variants and domain-general terms, 663 candidate pairs remained. Applying a clinical relevance threshold of 10 note-sections yielded 51 clinically valid transformation pairs accounting for 1,156 confirmed events. Confirmed substitutions represented approximately 1.2% of all consumer-term deletions, indicating that most editing involves higher-order discourse restructuring rather than direct one-to-one lexical replacement.

Transformation events were concentrated in narrative reasoning sections. The A&P accounted for 4,491 events (59.3%) and HPI for 2,981 events (39.3%), while Physical Exam (59; 0.8%) and Results (45; 0.6%) contributed minimally, consistent with their structured and terminology-stable roles. The disproportionate concentration in A&P, relative to its 34.2% share of note-sections, suggests that clinical reasoning and management planning contexts create conditions most conducive to targeted consumer-to-professional terminology normalization. Analysis of the full pair distribution revealed that clinically precise substitutions were systematically underrepresented in frequency rankings. High-frequency confirmed pairs such as therapy to treatment (n = 99) and medication to drug (n = 96) reflected broad professional framing, whereas diagnostically specific substitutions including high blood pressure to hypertension, degenerative arthritis to osteoarthritis, sugar to glucose, brittle bones to osteoporosis, and afib to atrial fibrillation occurred at lower frequencies (individual pair counts ranging from 1 to 20 note-sections). This frequency-validity inversion reflects natural variation in patient language, in which precise lay synonyms appear less often than general conversational descriptors. Domain-stratified inspection showed that confirmed substitutions spanned multiple clinical areas, including cardiovascular, musculoskeletal, metabolic, neurological, respiratory, and preventive care domains. Table 4 presents representative clinically stable consumer-to-clinical transformation pairs selected from the 51 pairs meeting the predefined section-level stability threshold. Beyond direct substitutions, additional lexical shifts provided broader context for editing behavior. Clinician revisions consistently increased documentation scaffolding and professional framing language, including patient, clinician, reviewed, clinic, and medical care. Conversely, conversational fillers and informal discourse markers including reports, like, well, and some showed systematic reductions following post-editing.

**Figure 2.** Example of dictionary-confirmed consumer-to-clinical transformation in the A&P section (deidentified synthetic text).

**Table 4.** Representative examples of dictionary-confirmed consumer-to-clinical transformation pairs across clinical domains.

| Domain | Consumer term | Clinical term | Sections (n) |
| --- | --- | --- | --- |
| General clinical | Doctor | Clinician | 58 |
| Medication | Pills | Medication | 18 |
| Cardiovascular | High blood pressure | Hypertension | 10 |
| Cardiovascular | Afib | Atrial fibrillation | 2 |
| Musculoskeletal | Degenerative arthritis | Osteoarthritis | 20 |
| Metabolic | Sugar | Glucose | 19 |
| Preventive care | Flu shot | Vaccine | 11 |
| Rheumatology | Rheumatoid arthritis | RA | 10 |
| Neurological | Seizure disorder | Epilepsy | 3 |
| Bone health | Brittle bones | Osteoporosis | 1 |

*Clinician-Level Variability*
Clinician-level analysis confirmed substantial between-clinician heterogeneity in editing intensity within each section type (Kruskal-Wallis consumer median $\chi^2$ = 17.04, p < 0.001; clinical median $\chi^2$ = 12.99, p = 0.005; Table 5). HPI exhibited the widest inter-clinician variability (IQR of clinician medians = 35.6 consumer terms), while Physical Exam (IQR = 3.8) and Results (IQR = 7.0) showed narrow, constrained distributions. The A&P section showed intermediate variability (IQR = 13.5). HPI differed significantly from Physical Exam after Holm correction (pairwise Mann-Whitney U, p = 0.002).

**Table 5.** Clinician-level editing heterogeneity by section. Changes represent draft-to-final change; negative values indicate reductions and positive values indicate increases.

| Section | Clinicians (N) | Median consumer-term change (IQR) | Median clinical-term change (IQR) | Median consumer removed (%) |
| --- | --- | --- | --- | --- |
| HPI | 90 | -8.0 (35.6) | -10.0 (39.3) | 71.4% |
| A&P | 61 | -4.5 (13.5) | -7.0 (21.5) | 65.7% |
| Results | 21 | -3.0 (7.0) | -4.0 (8.0) | 65.4% |
| Physical Exam | 22 | -2.0 (3.8) | -3.0 (6.4) | 60.4% |

Among 207 eligible clinicians, unsupervised clustering identified three editing profiles (Table 6). Most clinicians (n = 175; 84.5%) were moderate editors (mean consumer-term change = -16.5, clinical-term change = -19.2; 5.5% zero-change sections). A smaller cluster of high-intensity editors (n = 28; 13.5%) exhibited substantially larger changes (mean consumer-term change = -133.2, clinical-term change = -142.1; 1.7% zero-change), likely reflecting wholesale restructuring of AI drafts. A minimal-editing group (n = 4; 1.9%) accepted drafts with near-zero change (100% zero-change sections). Editing profile distribution did not differ significantly across specialty groups (Kruskal-Wallis p = 0.083), indicating that editing profiles were not driven by specialty norms.

**Table 6. Editing intensity profiles from unsupervised clustering.**

| Profile | n (%) | Mean clinical-term change | Zero-change rate |
|---|---|---|---|
| Moderate editors | 175 (84.5) | -19.2 | 5.5% |
| High-intensity editors | 28 (13.5) | -142.1 | 1.7% |
| Minimal editors | 4 (1.9) | 0.0 | 100.0% |

*Specialty-Stratified Editing Patterns*
Among the 214 classifiable clinicians with identifiable specialties, editing patterns showed consistent reductions across Medical specialties (median consumer-term change = -7.0, clinical-term change = -9.0), Primary Care specialties (-10.0, -11.0), and Surgical specialties (-13.0, -16.0), with Surgical specialties exhibiting the largest magnitude of change (Table 7). These differences were not statistically significant (Kruskal-Wallis p = 0.083), suggesting that specialty-level variation reflects differences in section-type distribution, particularly the proportion of narrative HPI content, rather than intrinsic specialty-specific editing behavior.

**Table 7.** Specialty-stratified editing patterns.

| Specialty Group | Clinicians (N) | Note-sections (n) | Median consumer-term change | Median clinical-term change | Mean consumer-term change | Mean clinical-term change |
|---|---|---|---|---|---|---|
| Medical Specialty | 84 | 12,123 | -7.0 | -9.0 | -17.1 | -19.2 |
| Primary Care | 57 | 14,610 | -10.0 | -11.0 | -28.9 | -30.7 |
| Surgical Specialty | 64 | 6,645 | -13.0 | -16.0 | -23.4 | -27.2 |
| Unknown | 9 | 1,191 | +5.0 | +7.0 | +4.5 | +6.6 |

**Discussion**
This study examined how clinicians transform conversational ambient AI draft notes into finalized clinical documentation within a real-world workflow. Across corpus, note, section, and clinician levels, post-editing was dominated by discourse-level restructuring such as deletion, consolidation, and professional framing, with selective insertion of clinical terminology. The consistent reduction in both consumer and clinical terminology density from draft to final documentation suggests that ambient AI drafts tend to over-generate lexical material and that clinicians primarily act as compressors and organizers of narrative content, not simple translators of lay expressions into professional synonyms. In this sense, post-editing appears to function as a normalization step that re-establishes clinical record conventions, evidentiary alignment, concise summarization, and standardized framing of clinical reasoning.

Section context emerged as the dominant determinant of editing intensity and confirmed transformations. Narrative sections, particularly A&P and HPI, exhibited the largest reductions and the widest variability in editing behavior, whereas Physical Exam and Results showed smaller, constrained changes consistent with their structured communicative roles. This pattern localizes where conversational drift is most costly. HPI drafts may capture patient phrasing and temporal detail that clinicians later compress into clinically salient signals, whereas A&P drafts are likely to require active reframing into professional reasoning and action-oriented plans. The disproportionate concentration of dictionary-confirmed substitutions in A&P, despite representing a smaller share of all note-sections than HPI, suggests that management planning contexts create frequent opportunities for clinicians to replace conversational descriptors with standardized professional terms as they formalize diagnoses, treatments, and follow-up language.

Dictionary-confirmed transformation analysis further clarifies what normalization means operationally in this workflow. Although 7,576 consumer-to-clinical transformation events were detected across 4,114 note-sections, these events represented only a small fraction of all consumer-term deletions. This discrepancy indicates that most clinician editing effort reflects higher-level discourse operations not captured by one-to-one lexical mapping: removing redundancy, collapsing narrative detail, reorganizing temporal sequences, and inserting professional scaffolding that signals clinical assessment, evidence review, and decision-making. The co-occurrence definition used here is intentionally conservative, requiring simultaneous removal of a consumer term and addition of its mapped clinical equivalent within the same note-section. Confirmed substitutions should therefore be interpreted as high-precision indicators of targeted lexical professionalization, embedded within a broader pattern of narrative standardization.

Clinician-level analyses revealed substantial heterogeneity in editing intensity that exceeded variation attributable to specialty classification. This suggests that ambient documentation interacts strongly with individual documentation style and risk tolerance. Some clinicians may systematically restructure drafts to match their preferred reasoning structure and phrasing, while others may accept drafts with minimal modification. The presence of a small minimal editing profile alongside a high-intensity cluster supports the interpretation that ambient AI does not simply shift time from typing to reviewing, it introduces a new spectrum of editing strategies, potentially shaped by workflow pressures, familiarity with the system, and preferences for how clinical reasoning is expressed. The lack of statistically significant differences across specialty groups is consistent with the idea that observed specialty differences are largely compositional, driven by section mix, especially the proportion of narrative HPI content.

These findings extend prior evaluations of ambient AI documentation, which primarily emphasize efficiency, clinician experience, and time savings[7,16-18]. The present results suggest a mechanism that may underlie those outcomes. Clinicians may achieve efficiency not by replacing writing with simple verification, but by performing rapid compression and professional framing of verbose drafts. Prior reports that describe post-editing as a review step may understate the linguistic work required to convert conversational narratives into clinically legible records[19,20]. From a design perspective, the dominant burden appears to reside in narrative structuring, not terminology mapping alone. Systems that optimize only lexical substitution or dictionary-based normalization are therefore unlikely to meaningfully reduce editing effort unless they also support section-aware summarization, redundancy control, and reasoning-aligned plan formulation.

The findings also refine expectations from consumer health informatics. CHV-based work highlights lexical gaps between lay and professional language[8,10], but real-world clinician editing appears to treat that gap as only one component of documentation quality. Clinicians' edits suggest that professionalization is not solely about vocabulary choice but also about evidentiary stance, attribution, and the organization of narrative into clinically interpretable structure. This distinction matters for building normalization tools. The most impactful improvements may require modeling section-specific discourse goals, such as patient-reported history or management decisions, while supporting clinicians in maintaining both clarity and concision.

The coexistence of reductions in conversational phrasing with selective insertion of professional terminology also reflects the dual-audience nature of contemporary documentation. As patients increasingly access notes through open-notes policies, clinicians may implicitly balance patient-legible phrasing with professional precision. In this context, normalization does not necessarily mean eliminating consumer language, it may mean strategically retaining understandable phrasing while ensuring that A&P content is expressed in standardized, clinically actionable terms. Evaluating how these editing patterns affect patient comprehension and trust is a consequential next step, particularly because compression and professional framing may improve clinical utility while simultaneously increasing linguistic distance from patient language.

Strengths of this study include analysis of a large real-world corpus of paired AI draft and finalized notes, integration of CHV resources, and section-level localization of editing behavior. Limitations include reliance on dictionary-based matching that may not capture semantic equivalence or contextual nuance, section aggregation dependent on structured export labels, inference of editing behavior from textual differences without direct measurement of clinician cognitive processes, and findings from a single academic health system that may not generalize broadly. Future work should examine whether substitution patterns cluster into clinically meaningful categories and evaluate how linguistic professionalization relates to documentation efficiency and patient comprehension.

**Conclusion**
Clinician post-editing of ambient AI-generated draft notes reflects structured linguistic normalization shaped by section context and individual editing style. Dictionary-confirmed consumer-to-clinical substitutions represent a small but reproducible subset of broader discourse restructuring, focused in narrative reasoning sections. These findings highlight the need for section-aware ambient AI design that supports clinician-driven professionalization while preserving workflow efficiency and documentation clarity.